%% file: egpaper_final.tex
\documentclass[10pt,twocolumn,letterpaper]{article}

\usepackage{3dv}
\usepackage{times}
\usepackage{epsfig}
\usepackage{graphicx}
\usepackage{amsmath}
\usepackage{amssymb}
\usepackage{wrapfig}
\usepackage[numbers,sort&compress]{natbib}
\usepackage{hyperref}

\threedvfinalcopy 

\def\threedvPaperID{184} 

\input{chapters/99_preamble.tex}

\ifthreedvfinal\fi
\newif\ifappendixonly
\appendixonlyfalse

\begin{document}

\title{A Spatio-temporal Transformer for 3D Human Motion Prediction}

\author{Emre Aksan\textsuperscript{1} \quad Manuel Kaufmann\textsuperscript{1} \quad Peng Cao\textsuperscript{2}\thanks{Affiliated with Peking University at the time of work.} \quad Otmar Hilliges\textsuperscript{1} \\ 
    \normalsize\textsuperscript{1}ETH Zürich, Department of Computer Science \quad 
    \textsuperscript{2}Massachusetts Institute of Technology\\
    {\tt\small \textsuperscript{1}\{eaksan,kamanuel,otmarh\}@inf.ethz.ch \textsuperscript{2}pengcao@mit.edu}
    }

    \setlength{\topsep}{0.5pt}
    \setlength{\parskip}{1ex}
    \renewcommand{\floatsep}{1ex}
    \renewcommand{\textfloatsep}{1ex}
    \renewcommand{\dblfloatsep}{1.5ex}
    \renewcommand{\dbltextfloatsep}{1.5ex}
    
\ifappendixonly
    \appendix
    \twocolumn[
    \centering
    \Large
    Supplementary Material for \\
    \vspace{0.5em}
    \textbf{A Spatio-temporal Transformer for 3D Human Motion Prediction} \\
    \vspace{1.0em}
    ]

    \setcounter{figure}{7}
    \setcounter{table}{3}
    \input{chapters/9_appendix.tex}

    \bibliographystyle{ieee_fullname}
    \bibliography{egbib}

\else

    \maketitle
    
    \begin{abstract}
        We propose a novel Transformer-based architecture for the task of generative modelling of 3D human motion. 
        Previous work commonly relies on RNN-based models considering shorter forecast horizons reaching a stationary and often implausible state quickly. 
        Recent studies show that implicit temporal representations in the frequency domain are also effective in making predictions for a predetermined horizon. 
        Our focus lies on learning spatio-temporal representations autoregressively and hence generation of plausible future developments over both short and long term.
        The proposed model learns high dimensional embeddings for skeletal joints and how to compose a temporally coherent pose via a decoupled \emph{temporal} and \emph{spatial} self-attention mechanism.
        Our dual attention concept allows the model to access current and past information directly and to capture both the structural and the temporal dependencies explicitly.
        We show empirically that this effectively learns the underlying motion dynamics and reduces error accumulation over time observed in auto-regressive models. Our model is able to make accurate short-term predictions and generate plausible motion sequences over long horizons. We make our code publicly available at \small{ \url{https://github.com/eth-ait/motion-transformer}}.
    \end{abstract}
    

    \input{chapters/1_introduction.tex}
    \input{chapters/2_related_work.tex}
    \input{chapters/3_method.tex}
    \input{chapters/4_experiments.tex}
    \input{chapters/8_conclusion.tex}
     \paragraph*{Acknowledgments}
     \vspace{-0.2cm}
     \begin{wrapfigure}{R}{0.45\columnwidth}
        \centering
        \hspace{-.5in}
        \vspace{-.1in}
        \includegraphics[width=0.45\columnwidth, trim=20 150 20 280]{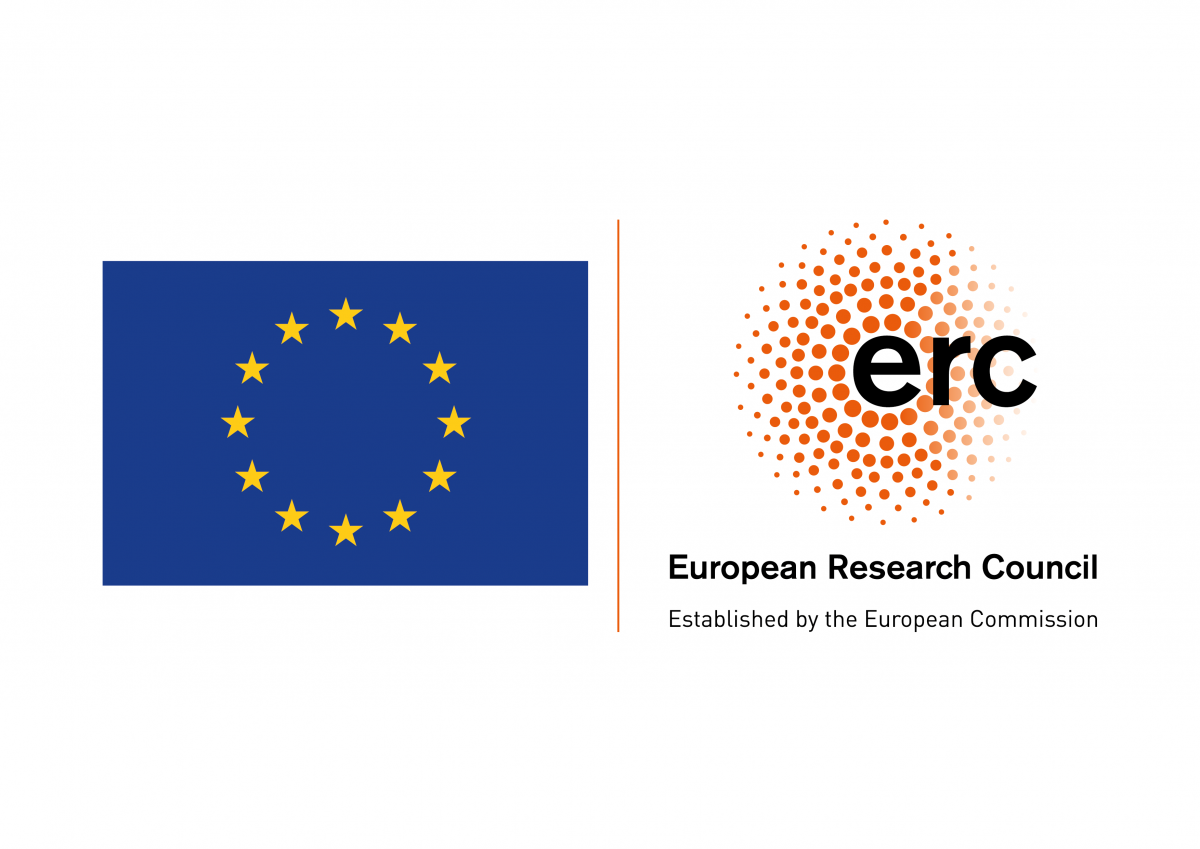}
         \end{wrapfigure}
        This project has received funding from the European Research Council (ERC) under the European Union's Horizon 2020 research \\ and innovation programme grant agreement No 717054.
    
    \clearpage
    {\small
    \bibliographystyle{ieee_fullname}
    \bibliography{egbib}
    }
    
    \clearpage
    
    \appendix
    \twocolumn[
    \centering
    \Large
    Supplementary Material for \\
    \vspace{0.5em}
    \textbf{A Spatio-temporal Transformer for 3D Human Motion Prediction} \\
    \vspace{1.0em}
    ]
    \input{chapters/9_appendix.tex}
\fi

\end{document}

%% file: chapters/99_preamble.tex
\usepackage[usenames,dvipsnames]{xcolor}

\usepackage{algorithmic}
\usepackage{algorithm}
\usepackage{rotating}
\usepackage{diagbox}
\usepackage{balance}

\usepackage[utf8]{inputenc} 
\usepackage[T1]{fontenc}    
\usepackage{url}            
\usepackage{booktabs}       
\usepackage{amsfonts}       
\usepackage{amsmath}
\usepackage{nicefrac}       
\usepackage{microtype}      
\usepackage{graphicx}       

\usepackage{paralist}
\usepackage{xspace}
\usepackage{comment}
\usepackage{subcaption}

\newcommand{\vectr}[1]{\boldsymbol{#1}}
\newcommand{\vectrT}[2]{\vectr{#1}_{#2}}
\newcommand{\vectrJ}[2]{\vectr{#1}^{(#2)}}
\newcommand{\vectrTJ}[3]{\vectrT{#1}{#2}^{(#3)}}

\newcommand{\myparagraph}[1]{\vspace{0.1em}\noindent\textbf{#1}}

\newcommand{\figref}[1]{Fig.~\ref{#1}}
\newcommand{\eqnref}[1]{Eq.~\ref{#1}}
\newcommand{\tabref}[1]{Tab.~\ref{#1}}
\newcommand{\secref}[1]{Sec.~\ref{#1}}
\newcommand{\norm}[1]{\left\lVert#1\right\rVert}



%% file: chapters/9_appendix.tex

The Supplementary Material includes this document and
a video. We provide additional details on the implementation of the proposed ST-Transformer and further experimental evidence of its performance.
We explain experimental details in \secref{sec:imp_details_appendix} and details of the multi-head attention mechanism in \secref{sec:mha_appendix}. We visualize more attention weights in \secref{sec:attention_viz_appendix} and \secref{sec:hyperparams_appendix} provides additional ablations.
In \secref{sec:more_2d_ablation} we provide more insights into how our ST-attention compares to naive 2D attention.
\secref{sec:ltd_appendix} and \secref{sec:ltd_attention_appendix} detail how we evaluated the LTD \cite{Mao_2019_ICCV} and LTD-Attention \cite{wei2020his} models on the AMASS dataset, respectively. Finally, in \secref{sec:ps_metrics_appendix} we provide definitions of the Power Spectrum metrics.

\begin{figure*}[t]
	\centering
	\includegraphics[width=0.9\textwidth]{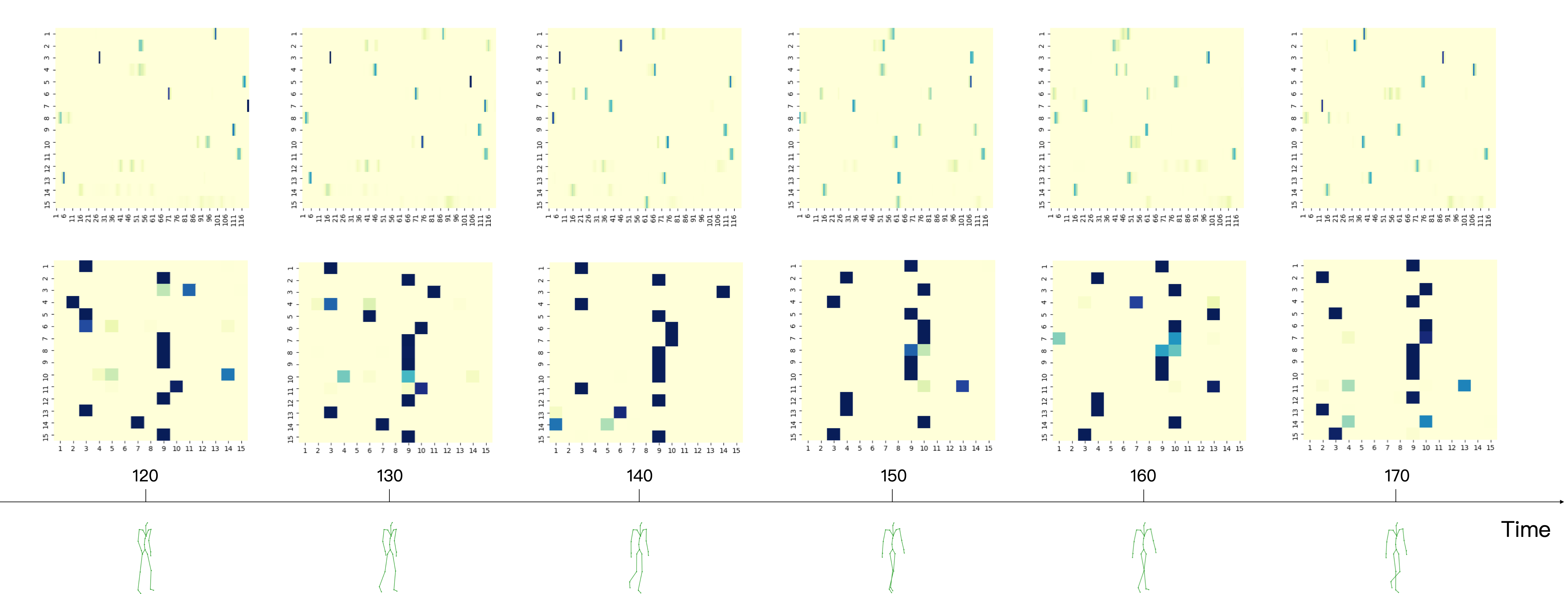}
	\caption{Temporal (\textit{top}) and spatial (\textit{middle}) attention weights aligned with the poses (\textit{bottom}) over $1$ second prediction. We visualize a single head in the first layer. Our model identifies the informative joints dynamically. Note that the temporal attention is calculated over a sliding window. Some of the past frames remain in the focus of the temporal attention despite the shift in the inputs. Similarly, the spatial focus is preserved on the more static joints such as left and right collars (i.e., columns $9$ and $10$ in the middle row).}
	\label{fig:attn_over_time}
\end{figure*}

\section{Experimental Details}
\label{sec:imp_details_appendix}

We implemented our models in TensorFlow \cite{tensorflow}. Hyper-parameters we used for the experiments are listed in Tab. \ref{tab:hyper_params}. Due to the limited data size of H3.6M, we achieved better results by using a smaller network as larger models usually suffered from overfitting.

As suggested by Vaswani \etal \cite{Vaswani2017Attention}, the Transformer architecture is sensitive to the learning rate. We apply the same learning rate schedule as proposed in \cite{Vaswani2017Attention}. The learning rate is calculated as a function of the training step as follows:
\begin{equation*}
\textit{learning rate} = D^{-0.5}\cdot \text{min}\left(\text{step}^{-0.5}, \text{step} \times \text{warmup}^{-1.5} \right),
\end{equation*}
where $D$ is the joint embedding size. $\text{Warmup}$ is set to $10000$ for our AMASS and H3.6M datasets. We use a batch size of $32$ and the Adam optimizer \cite{kingma2014adam} with its default parameters. Before updating the parameters, gradients are clipped with respect to the global gradient norm (i.e., clip by global norm) with a maximum norm value of $1.0$.

Following the training protocol in \cite{Aksan_2019_ICCV}, we apply early stopping with respect to the joint angle metric. 
Since our approach does not fall into the category of  sequence-to-sequence (seq2seq) models, we use the entire sequence (i.e., seed and target) for training. On H3.6M, the temporal attention window size is set to $75$ frames (i.e., $2$-sec seed and $1$-sec target at $25$ fps). On AMASS, we fed the model with sequences of $120$ frames ($2$-sec seed at $60$ fps) due to memory limitations. We followed an auto-regressive approach and train our model by predicting the next pose given the frames so far. In other words, the input sequence is shifted by $1$ step to obtain the target frames.

\begin{table}[b]
	\center
	\renewcommand{\arraystretch}{1.2}
	\setlength\tabcolsep{3pt}%
	\caption{H3.6M and AMASS experiment hyper-parameters.}
	\vspace{-0.5em}
	\begin{tabular}{l|c|c}
		& H3.6M & AMASS \\
		\hline
		Batch Size                  & 32    & 32    \\
		Window size (num. frames)               & 75    & 120   \\
		Num. Attention Blocks (L)     & 8     & 8     \\
		Num. Attention Heads (H)      & 4     & 8     \\
		Joint Embedding Size (D)    & 64    & 128   \\
		Feedforward Size 			& 128   & 256  \\
		Dropout Rate                & 0.1   & 0.1   \\
		\hline
	\end{tabular}
	\label{tab:hyper_params}
	\vspace{0.3cm}
\end{table}

\begin{table}[b]
	\center
	\renewcommand{\arraystretch}{1.2}
	\setlength\tabcolsep{3pt}%
	\caption{Our models performance when trained with different data representations. We report only the Euler error performance at $400$ms on the AMASS dataset.}
	\vspace{-0.3cm}
	\begin{tabular}{l|c}
		Representation & Euler Error \\
		\hline
		Quaternion & 0.543 \\
		6D Rotations & 0.527 \\
		Angle-axis & 0.522 \\
		Rotation Matrix & 0.490 \\
		\hline
	\end{tabular}
	\label{tab:data_formats}
\end{table}

Each attention block contains a feed forward network after the temporal and spatial attention layers. This feed forward network consists of two dense layers where the first one maps the $D=128$ dimensional joint embeddings into $256$-dimensional space for AMASS ($128$-dimensional space for H3.6M), followed by a ReLU activation function. The second dense layer always projects back into the $D$-dimensional joint embedding space. We use the same dropout rate of $0.1$ for all dropout layers in our network.

\noindent \textbf{Data represenations} $3$D joint angles can be represented with various representations such as angle-axis \cite{Martinez2017Motion}, quaternion \cite{Pavllo2018} or rotation matrix \cite{Aksan_2019_ICCV}. In our experiments both on the AMASS and H3.6M datasets, we trained and evaluated our model on all three joint angle representations. Similarly, we also evaluated the baseline models LTD and LTD-Attention (cf. \secref{sec:ltd_appendix} and \secref{sec:ltd_attention_appendix}) by using all the joint angle representations and reported the best performance.

We found that all models achieved their best performances with the rotation matrix representation (see \tabref{tab:data_formats} for our model's results).
Since the model's predictions might not be valid rotation matrices, we project them to the nearest valid rotation matrix in SO(3) using the singular value decomposition.
The evaluation metrics are then computed on the projected rotation matrices.

\noindent\textbf{Weight sharing}
In our initial design both temporal and spatial attention were alike (i.e., following Eq.~(3)).
We experimentally found that our current design, i.e., sharing the \textit{key} and \textit{value} weights across joints but keeping the \textit{query} separate, achieves better performance. We keep the temporal attention as is and compare our current design (i.e., Eq.~(5)) with two alternatives: projection weights \textbf{$W^Q$}, \textbf{$W^K$}, \textbf{$W^V$} are (A) separate (i.e., joint-wise as in Eq.~(3)) or (B) shared across joints. On AMASS at $400$ms, (A) achieves $0.511$ Euler error and (B) $0.504$ (vs $0.490$ with the current design).

\noindent \textbf{Data augmentation} Finally, we observed a benefit of data augmentation on the H3.6M dataset. With a random chance of 0.5, we reverted or mirrored a sample sequence. Mirroring the joints was previously reported to be useful in \cite{Pavllo2018}. For the former one, we hypothesize a reverted sequence still posses valid human poses and backward motion dynamics, which may help the model to minimize the null space.  

\begin{figure}[b]
	\centering
	\includegraphics[width=\columnwidth]{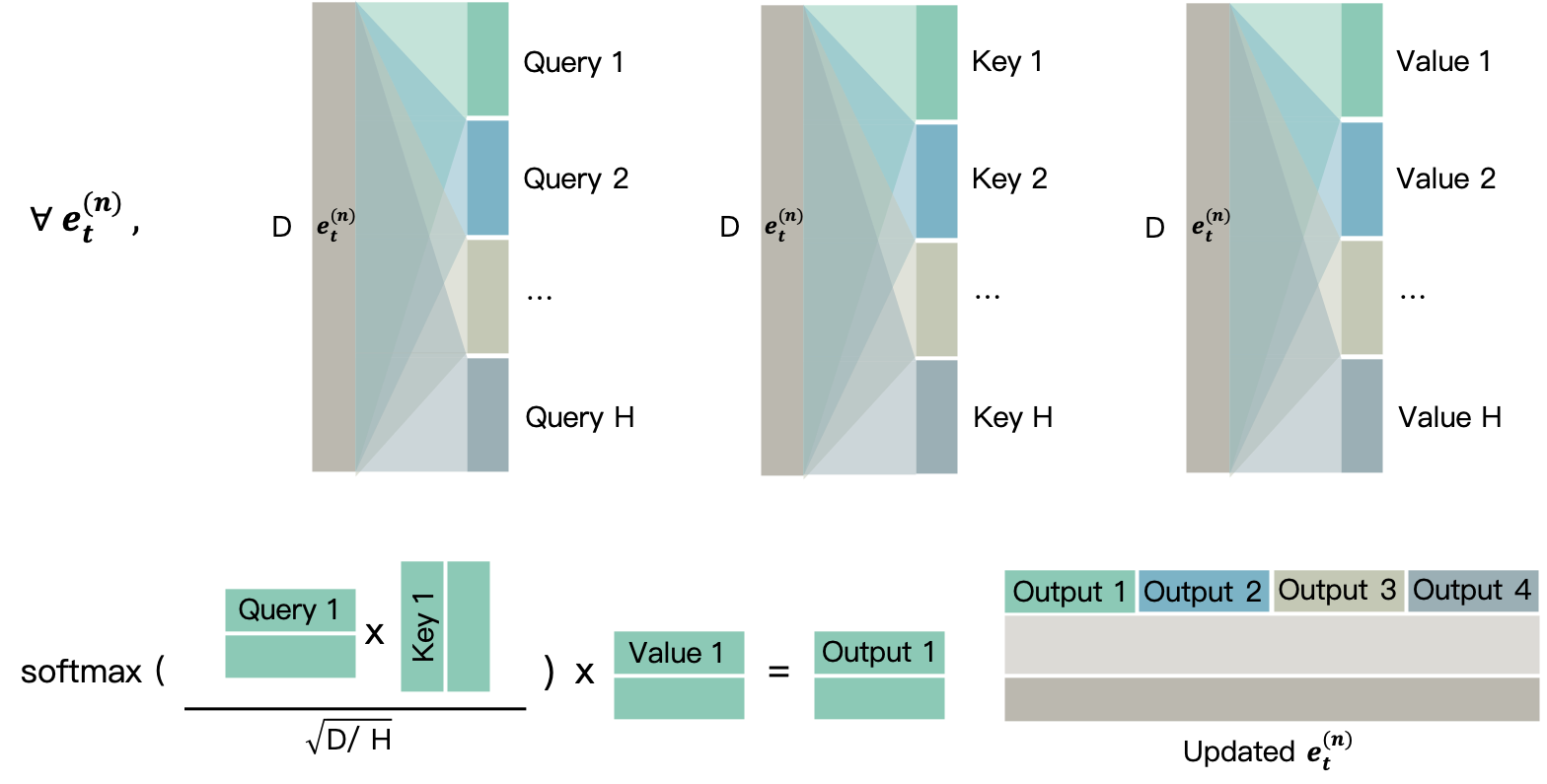}
	\caption{The multi-head self-attention mechanism. Each color represents one head.}
	\label{fig:multi_attn}
\end{figure}

\section{Multi-head Attention}
\label{sec:mha_appendix}
\figref{fig:multi_attn} illustrates the multi-head self-attention mechanism mentioned in Eq. (3) and (4) of the main paper. The sub-spaces of query, key, and value for each attention head are calculated from the joint embeddings $\vectrTJ{e}{t}{n}$. From this figure it becomes clear that the joint embedding size $D$ must be evenly divisible by the number of heads $H$. This in turn means the projection matrices involved for query, key, and value are of dimension $\mathbb{R}^{D \times \ell}$, where $\ell=D/H$. In the main paper we refer to $\ell$ as either $S$ or $F$ to distinguish between the temporal and spatial attention layer.

In the temporal attention blocks, we use separate query, key, and value weight matrices for different joints. In the spatial attention blocks, while the key and value weight matrices are shared across joints, the query weight matrices are not. Having obtained all the query, key, and value embeddings, we can get the output of each head according to Eq. (4) in the main submission. Note that the attention is over $T$ time steps in the temporal attention blocks and $N$ joints in the spatial attention blocks.
Finally, the outputs of all heads are concatenated and then fed to a feed forward network consisting of two dense layers and computing the updated embedding of $\vectrJ{\Bar{E}}{n}$.

\section{What does Attention Look Like?}
\label{sec:attention_viz_appendix}
In order to adapt to changing spatio-temporal patterns, our model calculates the attention weights at every step. In \figref{fig:attn_over_time}, we visualize the change in the attention weights over time. The focus of the network changes as expected. Although the attention window shifts in time, we observe that for quasi-static joints like the hips or spine, the model maintains the focus on the same time-step in this particular attention head. When we calculate the inter-joint dependencies via spatial attention, a large number of joints attend to the left and right collars. This could indicate that such mostly static joints are used as reference.

\section{Hyper-parameters}
\label{sec:hyperparams_appendix}
\begin{figure}[h]
	\centering
	\begin{subfigure}[t]{\columnwidth}
		\centering
		\includegraphics[width=0.95\columnwidth]{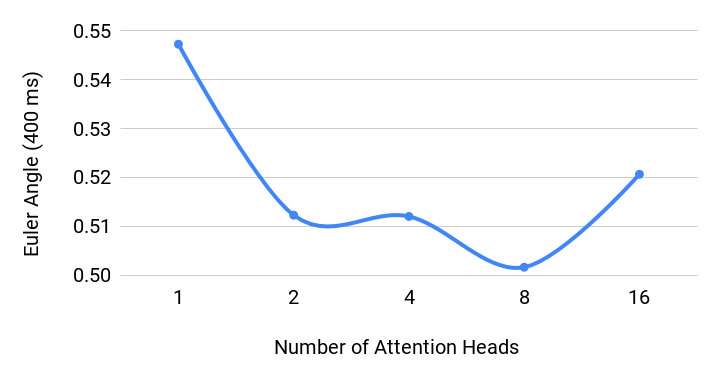}
		\caption{\small Different number of attention heads ($H$).}
		\label{fig:ablation_nheads}
	\end{subfigure}
	\begin{subfigure}[t]{\columnwidth}
		\centering
		\includegraphics[width=0.95\columnwidth]{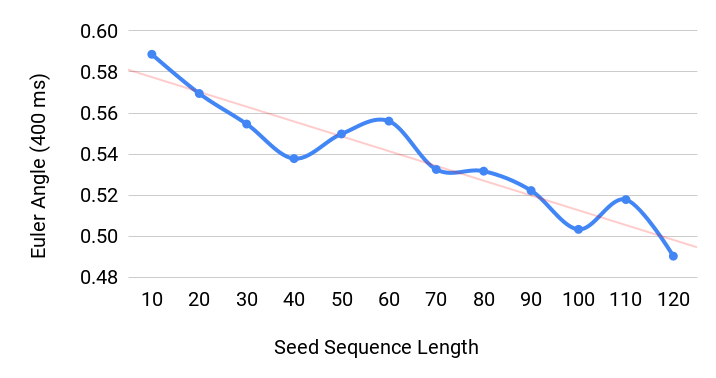}
		\caption{\small Seed sequences of different lengths.}
		\label{fig:ablation_seed_len}
	\end{subfigure}
	\begin{subfigure}[t]{\columnwidth}
		\centering
		\includegraphics[width=0.95\columnwidth]{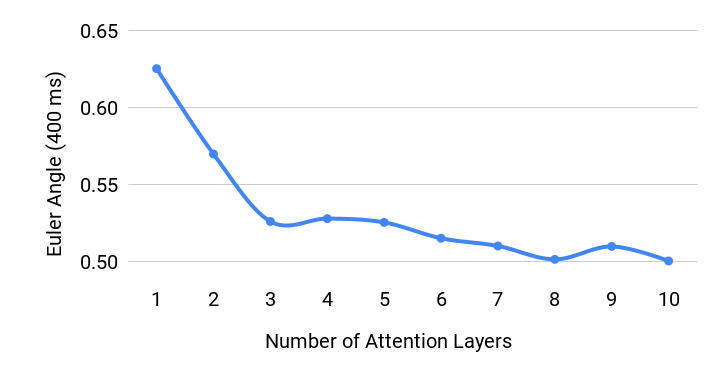}
		\caption{\small Different number of attention layers ($L$).}
		\label{fig:ablation_n_att_layers}
	\end{subfigure}
	\vspace{-0.1em}
	\caption{Ablations on the AMASS dataset. We evaluate our model with varying configurations and report Euler angle loss at $400$ ms.}
	\label{fig:mha}
\end{figure}

We experiment with varying number of attention heads $H$. As shown in \figref{fig:ablation_nheads}, the best performance on AMASS is achieved with $8$ attention heads. A model with $2$ attention heads also yields reasonable performance. Comparison between the performance of multi- and single-head attention mechanism suggests that the model benefits from using more than one spatio-temporal configuration.

\figref{fig:ablation_seed_len} plots the performance of our model when trained with seed sequences of varying length, showing the performance w.r.t. the temporal attention window. The decreasing trend suggests that our model benefits from longer sequences. This hypothesis is supported via the temporal attention masks showing that our model accesses poses from the beginning of the sequence (cf. \figref{fig:attn_over_time}).

We train our model with varying number of layers. \figref{fig:ablation_n_att_layers} shows that reasonable performance on AMASS is reached with only $3$ layers. However, as the number of layer increases, the representations learned by our model is tuned better. It can also be considered as the number of message passing steps to update the available representation.

\section{Additional Ablation on 2D Attention}
\label{sec:more_2d_ablation}

To provide more insights into the computational efficiency of our ST-attention compared to the naive 2D attention discussed in Sec.~4.3, we present several additional comparisons in \tabref{tab:2d_ablation}. 

In rows 1-3 of \tabref{tab:2d_ablation}, we maximize one of the three hyper-parameters for the 2D attention where row 1 exhausts the batch size and row 2 the window size at the cost of the batch size.
We observe that prioritizing one of the three parameters is usually detrimental for the performance of the 2D attention model and we get the best result with a trade-off between the three (cf. rows 4, 6). Furthermore, we also observe that those best 2D configurations are always outperformed when switching to our decoupled ST attention (cf. rows 5, 7). Row 8 reports the best configuration we found with our ST-attention.
These results show that our decoupled attention mechanism is superior to the plain 2D attention even when controlling for computational efficiency.

\begin{table}[t]
    \centering
    \renewcommand{\arraystretch}{1.3}
	\setlength\tabcolsep{4pt}%
    \resizebox{\linewidth}{!}{%
    \begin{tabular} {c l c c | c c | c c }
		& & \multicolumn{2}{c}{Euler $\downarrow$} & \multicolumn{2}{c}{Joint Angle $\downarrow$} & \multicolumn{2}{c}{Positional $\downarrow$} \\       
		ID & Milliseconds $\rightarrow$ & 100 & 400 & 100 & 400 & 100 & 400 \\
		\hline
		1 & {[}2D{]} L4 - W80 - B32           & 0.218 & 0.583 & 0.038 & 0.121 & 14.4 & 46.9 \\
        2 &  {[}2D{]} L8 - W120 - B5           & 0.256 & 0.627 & 0.044 & 0.129 & 16.5 & 50.2 \\
        3 &  {[}2D{]} L8 - W60 - B16           & 0.238 & 0.602 & 0.041 & 0.123 & 15.7 & 48.6 \\
        \hline
        4 &  {[}2D{]} L4 - W100 - B16 & 0.213       & 0.555    & 0.037      & 0.115       & 14.5        & 45.2       \\
        5 & {{[}ST{]} L4 - W100 - B16} & {0.201}       & {0.538}       & {0.035}       & {0.111}       & {13.1}        & {42.6}        \\
        \hline
        6 &  {{[}2D{]} L8 - W40 - B32}  & {0.212}       & {0.575}       & {0.036}       & {0.119}       & {13.6}        & {45.5}        \\
        7 &  {{[}ST{]} L8 - W40 - B32}  & {0.204}       & {0.538}       & {0.036}       & {0.111}       & {14.2}        & {43.6}        \\
        \hline
        8 &  \textbf{{[}ST{]} L8 - W120 - B32} & \textbf{0.178}       & \textbf{0.490}       & \textbf{0.033}       & \textbf{0.103}       & \textbf{12.8}        & \textbf{39.5}  \\
		\hline
	\end{tabular}
    }
     \caption{2D attention ablation. ST refers to our decoupled attention. Each row corresponds to a different configuration denoted by \textbf{L} (number of stacked attention layers), \textbf{W} (number of context frames, i.e., temporal attention length) and \textbf{B} (batch size).
     }
    \label{tab:2d_ablation}
\end{table}

\section{Evaluation of LTD on AMASS}
\label{sec:ltd_appendix}
\label{sec:app_dct_evaluation}
In Tab. 1 of the main paper we report the performance of the LTD model \cite{Mao_2019_ICCV} (cf. LTD-10-10 entry) on the AMASS dataset. The results correspond to the best results we obtained after hyper-parameter-tuning, explained in more detail in the following.

To train and evaluate LTD on AMASS we use the code provided by Mao \etal \cite{Mao_2019_ICCV} but swap out the data pipeline to load AMASS instead of H3.6M. In \cite{Mao_2019_ICCV} the inputs to the network and the outputs are 400 milliseconds (10 frames) worth of data. As AMASS is sample at 60 Hz, we hence pass 24 frames as input and let the model predict 24 frames. %

We fine-tune the learning rate as well as the number of DCT coefficients. For the number of DCT coefficients we use the original 35, but also the maximum number of coefficients 48.
As reported by \cite{Mao_2019_ICCV} we find this to make little difference, but 35 coefficients led to slightly better results.

The remaining hyper-parameters are as follows, which mostly corresponds the original setting. We are employing the Adam \cite{kingma2014adam} optimizer with a learning rate of 0.001 and batch size of 16. We train for maximum 100 epochs with early stopping. The learning rate is decayed by a factor of 0.96 every other epoch. 
The input window size is 48. The model parameters are kept as proposed in the original paper resulting in a model size of roughly $2.23$ Mio. parameters.

\section{Evaluation of LTD-Attention on AMASS}
\label{sec:ltd_attention_appendix}
To evaluate the LTD-Attention \cite{wei2020his} model on AMASS, we again use the publicly available code and plug in our AMASS data loading pipeline.
We adjust the number of input and output frames to reflect the different framerate on AMASS, i.e. we feed seeds of length $120$ ($2$ seconds) and predict $60$ frames ($1$ second). We have found that this resulted in better performance than predicting $24$ frames ($400$ milliseconds) directly.

We keep the Adam optimizer with the originally proposed learning rate decay, but fine-tuned the learning rate to $0.005$ with a batch size of 128. Similarly, we found $45$ DCT coefficients to work best and we kept the kernel size at the original value of $10$. The model is trained for $100$ epochs and all other model parameters are kept the same with the exception of the size of the inputs and outputs.
Like for our model, we also tried out different joint angle representations, of which rotation matrices performed best.

\section{Power Spectrum Metrics}
\label{sec:ps_metrics_appendix}
On AMASS, we show our model's capability in making very long predictions (i.e., up to $15-20$ sec). Such long prediction horizons prevent us from using pairwise metrics such as the MSE because the ground-truth targets are often much shorter than the prediction horizon. Hence, we use Power Spectrum (PS) metrics originally proposed by Hernandez \etal \cite{Hernandez_2019_ICCV}.

In order to adapt the metrics into our new setup, we slightly modify the evaluation protocol. We use 3D joint positions instead of angles as it is straightforward to convert any angle-based representation into positions by applying forward kinematics. This also allow us to compare models operating on arbitrary rotation representations. Given a sequence $\vectr{X}$, we treat every coordinate of every joint over time as a feature sequence $\vectr{x}_f$ following \cite{Hernandez_2019_ICCV}. The power spectrum PS is then equal to $PS(\vectr{x}_f) = ||FFT(\vectr{x}_f)||^2$ where $FFT$ denotes the Fast Fourier Transform.

\paragraph{PS Entropy} It is defined as
\begin{align*}
& PS~Entropy(\mathcal{X}) = \\ &\frac{1}{|\mathcal{X}|}\sum_{\vectr{X} \in \mathcal{X}} \frac{1}{F}\sum_{f = 1}^F\sum_{e=1}^E &-||PS(\vectr{x}_f)|| * 
\log (||PS(\vectr{x}_f)||)
\end{align*}

where $\mathcal{X}$ is either the ground-truth test or training dataset, or the predictions made of a respective model on the corresponding test dataset. $f$ and $e$ correspond to a feature and frequency, respectively.

\paragraph{PS KLD} To compute the PS KLD metric, we use the following approach. Instead of using the variable-length ground-truth targets, we randomly get $20'000$ sequences of length $1$ sec (i.e., $60$ frames) from the test dataset and calculate the power spectrum distribution $G$. Then, we get non-overlapping windows of $1$ sec from the predictions to get $P_{t}$ where $t$ stands for the corresponding prediction window of length 1 second. For example, $P_{5}$ is the power spectrum distribution for the predictions between $5$ and $6$ seconds. This enables us to measure the quality of arbitrarily long predictions by comparing every second of the predictions with the real reference data.

The symmetric PS KLD metric is then defined as
\begin{align*}
& PS~KLD(G, \mathcal{X}, t) = \\ 
&\frac{1}{2|\mathcal{X}|} \sum_{P_t \in \mathcal{X}} KLD(G \mid\mid P_t) + KLD(P_t \mid\mid G)
\label{eqn:sym_ps_kld}
\end{align*}

We use publicly available implementations of the metrics (\cite{Hernandez_2019_ICCV}, \href{https://github.com/magnux/MotionGAN/blob/master/test.py\#L634}{Github link}). It is worthwhile to mention that the PS KLD metric does not make pairwise comparisons between ground-truth and predictions. Instead, it measures the discrepancy between the real and predicted data distributions.

%% file: chapters/1_introduction.tex
\section{Introduction}
3D human motion modelling is typically formulated as the prediction of future poses given a past horizon. 
Humans are able to effortlessly forecast the complex dynamics of motion in a plausible fashion due to our strong structural and temporal priors.
From a learning perspective this problem can be seen as a generative modelling task: A network learns to synthesize a sequence of human poses, where the model is conditioned on the seed sequence.
The task requires learning of pose priors for natural articulation and of underlying dynamics to yield  plausible motion predictions.
Since these factors are highly latent and entangled, introducing inductive biases and tailoring architectures for the task is essential for modelling of 3D human motion data.

Given the temporal nature of human motion, it is not surprising that recurrent neural networks (RNNs) are the most popular choice  \cite{Aksan_2019_ICCV,Fragkiadaki2015ERD,Jain2016, Martinez2017Motion, Pavllo2018, Wang2018Adversarial}. 
RNNs model short and long-term dependencies by propagating information through their hidden state. 
Convolutional neural networks (CNN) in a sequence-to-sequence framework have also been proposed
\cite{Li_2018_CVPR, Hernandez_2019_ICCV, kaufmann2020convolutional}. 
Such approaches focus on modelling the temporal aspect of the problem following an auto-regressive approach, but neglect structural priors.
Instead, vectorized poses are passed as inputs at every step and the spatial dependencies are assumed to be learned implicitly. However, considering the skeletal structure in the architectural level is shown to be an effective inductive bias in \cite{Jain2016, Butepage2017RepL, li2019efficient, Aksan_2019_ICCV}.

Since the auto-regressive approach factorizes the predictions into step-wise conditionals based on previous predictions, these models tend to accumulate error over time and eventually the predictions collapse to a non-plausible pose. 
This issue can be associated with the exposure bias problem \cite{ranzato2015sequence} due to discrepancies between data and model distributions. 
Previous work has applied various strategies to work around this problem, such as using model predictions during training \cite{Martinez2017Motion, Pavllo2018}, applying noise to the inputs \cite{Aksan_2019_ICCV,Jain2016, Li_2018_CVPR,Ghosh2017}, or using adversarial losses \cite{Wang2018Adversarial, Li_2018_CVPR}. 

Recent works \cite{Mao_2019_ICCV, wei2020his, cai2020learning} model the temporal aspects of 3D human motion by encoding every joint's trajectory with the discrete cosine transformation (DCT). Both the observations and the predicted future frames are represented as a set of DCT coefficients which are then used to model inter-joint dependencies. Such an implicit modelling of the temporal information inherently mitigates the failure cases of the auto-regressive models. DCT appears to be an effective non-learning based representation. 

In this work, we present a novel architecture for 3D human motion modelling, which attempts to learn a spatio-temporal representation explicitly without relying on the propagation of a hidden state as in RNNs or fixed temporal encodings such as DCT coefficients. Our approach is motivated by the recent success of the Transformer model \cite{Vaswani2017Attention} in tasks such as NLP \cite{Vaswani2017Attention,devlin2018bert}, music \cite{huang2018music}, or images \cite{child2019generating, parmar2018image}. While the vanilla Transformer is designed for 1D sequences with a self-attention mechanism \cite{Vaswani2017Attention, al2019character, parikh2016decomposable}, we note that the task of 3D motion prediction is inherently spatio-temporal and we propose a novel representation that decouples the temporal and spatial dimensions.  

\begin{figure}[t]
	\centering
	\includegraphics[width=0.46\textwidth, trim={0pt 390pt 580pt 0pt}]{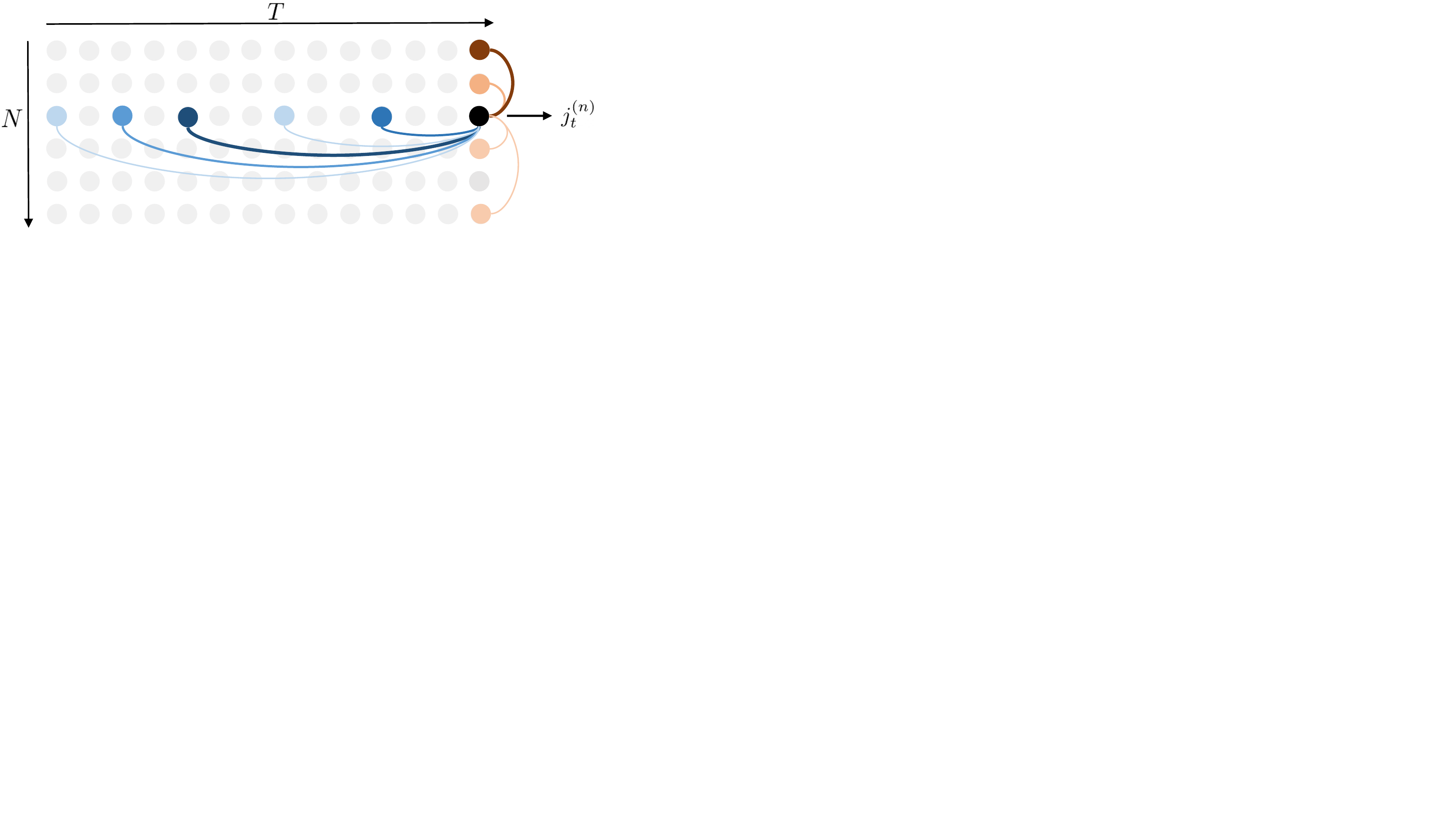}
  \caption{ \textbf{Spatio-temporal attention.} The motion sequence is a matrix containing $N$ joint configurations $\vectrTJ{j}{t}{n}$ over $T$ time steps. For a given joint (black), the \textit{temporal} (blue) and \textit{spatial attention} (red) are illustrated. Color intensity indicates attention weight.}
  \label{fig:attention2d}
\end{figure}

The proposed spatio-temporal attention mechanism is trained to identify useful information from a known sequence to construct the next output pose.
For every joint we define \textit{temporal attention} over the same joint in the past and \textit{spatial attention} over the other joints at the same time step (see \figref{fig:attention2d}).
The spatial attention block draws information from the joint features at the \textit{current} time step whereas the temporal block focuses on distilling information from the \textit{previous} time steps of individual joints.
A prediction is then made by summarizing current joint information and previous time steps as a weighted combination. 

Our model learns to construct temporally coherent poses from individual joints by considering the temporal and spatial representations learned from the data.
The dual self-attention over the sequence allows the model to access past information directly and hence capture the dependencies explicitly \cite{parikh2016decomposable, Vaswani2017Attention}, mitigating error accumulation over time. 
It also enables interpretability since attention weights indicate informative sequence parts that led to the prediction.
Our experiments show that a naive application of 1D self-attention still suffers from the collapsing pose problem, whereas our proposed model, the ST-Transformer, is able to outperform the state-of-the-art models in short-term horizons and also produce convincing long-term predictions (up to 20 seconds for periodic motions).

%% file: chapters/2_related_work.tex
\section{Related Work}
\label{sec:related_work}

Decomposition of the attention mechanism across different dimensions has been shown to be effective in other domains.
In \cite{huang2019ccnet}, attention is applied separately to the height and width dimensions of images for semantic segmentation. \cite{shi2020decoupled} proposes a similar model for skeleton-based action recognition where attention is applied sequentially first to the joints and then to the temporal dimension. \cite{ho2019axial} introduces axial attention applying self-attention on different dimensions in parallel to reduce computational complexity. Our work is conceptually similar to those but crucially differs in the task domain. In the remainder of this section, we provide a summary of the related work on 3D motion modelling.

\noindent \textbf{Recurrent Models}
RNNs are the dominant architecture for 3D motion modelling tasks \cite{Aksan_2019_ICCV,Fragkiadaki2015ERD,Jain2016,Ghosh2017,Chiu2018ActionAgnosticHP,Du2019Pedestrian,Wang_2019_ICCV}.
Fragkiadaki \etal \cite{Fragkiadaki2015ERD} propose the Encoder-Recurrent-Decoder (ERD) model where an LSTM cell operates in latent space.
Jain \etal \cite{Jain2016} build a skeleton-like st-graph with RNNs as nodes. 
Aksan \etal \cite{Aksan_2019_ICCV} replace the dense output layer of a RNN architecture with a structured prediction layer that follows the kinematic chain. The authors furthermore introduce a large-scale human motion dataset, AMASS \cite{AMASS2019}, to the task of motion prediction.
The error accumulation problem is typically combated by exposing the model to dropout or Gaussian noise during training. Ghosh \etal \cite{Ghosh2017} more explicitly train a separate de-noising autoencoder that refines the noisy RNN predictions.

Martinez \etal \cite{Martinez2017Motion} introduce a sequence-to-sequence (seq2seq) architecture with a skip connection from the in-~to the output on the decoder to address the transition problems between the seed and predictions. They also propose training the model with the predictions to alleviate the exposure bias problem. Similarly, Pavllo \etal \cite{Pavllo2018} suggest to use a teacher-forcing ratio to gradually expose the model to its own predictions. In \cite{Chiu2018ActionAgnosticHP}, the seq2seq framework is modified to explicitly model different time scales using a hierarchy of RNNs. Gui \etal \cite{Wang2018Adversarial} propose a geodesic loss and adversarial training and Wang \etal \cite{Wang_2019_ICCV} replace the likelihood objective with a policy gradient method via imitation learning. The acRNN of 
Zhou \etal \cite{zhou2018acRNN} uses an augmented conditioning scheme which allows for long-term motion synthesis. However their model is trained on specific motion types, whereas in this work we present a single model encompassing multiple motion types.

While the use of domain-specific priors and objectives can improve short term accuracy, the inherent problems of the underlying RNN architectures still exist. Maintaining long-term dependencies is an issue due to the need of summarizing the entire history in a hidden state of fixed size.
Inspired by similar observations in the field of NLP \cite{Vaswani2017Attention,parikh2016decomposable}, we introduce a spatio-temporal self-attention mechanism to mitigate this problem. In doing so we let the network explicitly reason about past frames without the need to compress the past into a single hidden vector.

\myparagraph{Non-recurrent Models}
Bütepage \etal \cite{Butepage2017RepL, Butepage2018Anticipating} use dense layers on sliding windows of the motion sequences. In \cite{Holden2016,Holden2015} convolutional models are introduced for motion synthesis conditioned on trajectories. More recently, Hernandez \etal \cite{Hernandez_2019_ICCV} propose to treat motion prediction as an image inpainting task and use a convolutional model with adversarial losses. Joints are represented as 3D positions, often requiring auxiliary losses such as bone length and joint limits to ensure anatomical consistency. Li \etal \cite{Li_2018_CVPR} use CNNs instead of RNNs in the sequence-to-sequence framework to improve long-term dependencies. Similarly, Kaufmann \etal \cite{kaufmann2020convolutional} propose a convolutional autoencoder for the 3D motion infilling task to fill in large gaps between given sequences.

\myparagraph{Implicit Temporal Models}
Mao \etal \cite{Mao_2019_ICCV} represent sequences of joints via discrete cosine transform (DCT) coefficients and train a graph convolutional network (GCN) to learn inter-joint dependencies. Since the GCN operates on temporal windows of poses and produces the entire output in one go, the predictions are limited to a pre-determined length. In follow-up work~\cite{wei2020his}, DCT coefficients are instead extracted from shorter sub-sequences
in an overlapping sliding window fashion which are then aggregated via a 1D attention block. Similarly, Cai \etal \cite{cai2020learning} leverage a Transformer architecture on the DCT coefficients extracted from the seed sequence and make joint predictions progressively by following the kinematic tree.

Our model is related to these approaches, but differs in three aspects. 
First, the DCT requires windowed inputs and produces the entire output in one go.
This limits full generative modelling of arbitrarily long sequences with sufficient diversity. Instead, we aim to learn spatio-temporal representations directly from the data. 
Second, we follow a fully auto-regressive approach and model the temporal dependencies explicitly by leveraging the recursive nature of human motion. 
Third, temporal and spatial modelling is interleaved in our design, whereas in previous work the temporal information is modeled first via the DCT \cite{Mao_2019_ICCV} and aggregated with an attention mechanism \cite{wei2020his, cai2020learning}, and then the spatial structure is captured by a GCN or a Transformer. In contrast, our model stacks several computation blocks each of which aggregates temporal and spatial information and passes it to the subsequent layer in a message passing fashion.

In summary, existing 3D motion modelling works have introduced regularization, structural priors, frequency transformations, or auxiliary and adversarial loss terms to address the inherent problems of the underlying architectures.
We show that the self-attention concept itself is very effective in learning motion dynamics and allows for the design of a versatile mechanism that is effective and easy to train.

%% file: chapters/3_method.tex
\section{Method}
\label{sec:method}

\begin{figure*}[t]
    \centering
    \includegraphics[width=0.95\textwidth]{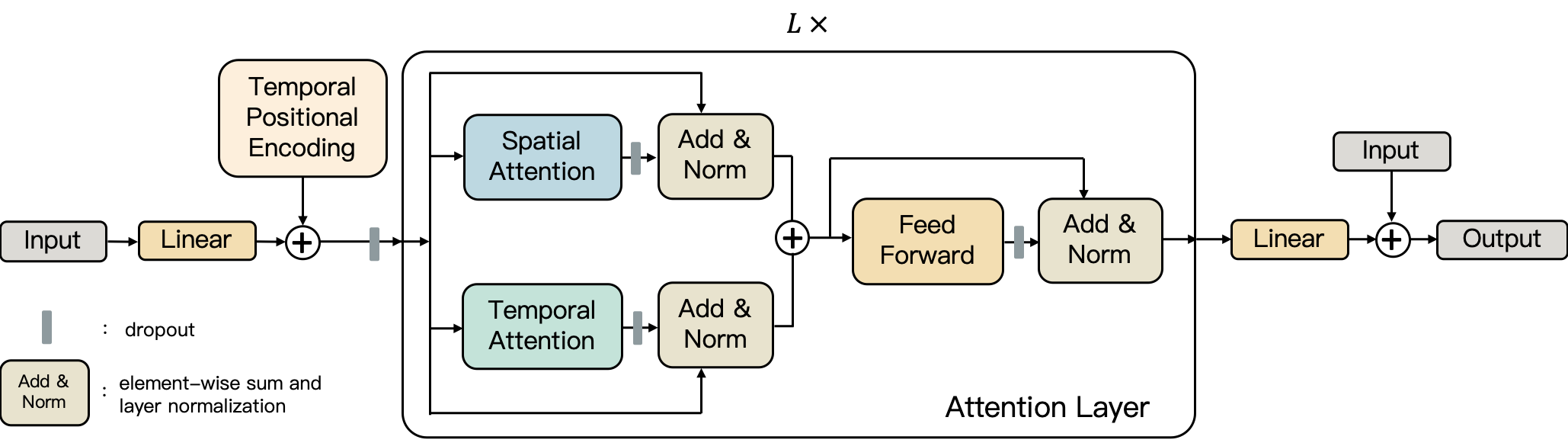}
    \caption{\textbf{Architecture overview.} We first project every joint into an embedding space and then inject positional encodings and apply dropout. Next, the embeddings are fed to $L=8$ stacked attention layers. We employ a novel spatio-temporal multi-head attention mechanism based on \cite{Vaswani2017Attention}. It is split into a \textit{temporal} attention block that updates a joint's embedding by looking at the past instances of the same joint and a \textit{spatial} block that attends over all the joints in the current time step. Finally, we estimate the next pose by projecting the embeddings back to the joint space and using a residual connection from input to output, following \cite{Martinez2017Motion}.
    }
    \label{fig:overview}
\end{figure*}

We now explain the architecture of the proposed spatio-temporal transformer (ST-Transfomer) in detail. For an overview please refer to \figref{fig:overview}.
Our method uses the building blocks of the Transformer \cite{Vaswani2017Attention}, but with two main differences: (1) a decoupled spatio-temporal attention mechanism and (2) a fully auto-regressive model.

\subsection{Problem Formulation}

A motion sample can be represented by a sequence $\vectr{X} =\{ \vectrT{x}{1}, \dotsc, \vectrT{x}{T}\}$ where a frame $\vectrT{x}{t} = \{\vectrTJ{j}{t}{1}, \dotsc, \vectrTJ{j}{t}{N}\}$ denotes a pose at time step $t$ with joints $\vectrTJ{j}{t}{n} \in \mathbb{R}^M$. Each joint $\vectr{j}$ is an $M$-dimensional vector where $M$ is determined by the pose parameterization, e.g., 3D position, rotation matrix, angle-axis, or quaternion. We use a rotation matrix representation, i.e., $M = 9$.
Following a predefined order, a sequence sample $\vectr{X}$ can be written as a matrix of size $NM \times T$, where blocks of $M$ rows represent the $i$-th joint configuration at time step $t$, i.e. $\vectr{X}_{Mi:M(i+1), \ t} = \vectrTJ{j}{t}{i}$.

In our notation the subscript denotes the time step. We use $n-$tuples in the superscript ordered by joint index, layer index, and optionally attention head index. For example, $\vectrJ{W}{n, I}$ denotes the weight matrix of the input ($I$) of joint $n$. Note that projection matrices $\vectr{W}$ and biases $\vectr{b}$ are trainable. The superscript $n$ indicates that it is only used by joint $n$.

\subsection{Spatio-temporal Transformer}
\myparagraph{\bf{Joint Embeddings}} We first project all joints into a $D$-dimensional space via a single linear layer, i.e.
$
\vectrTJ{e}{t}{n} = \vectrJ{W}{n, E} \vectrTJ{j}{t}{n} + \vectrJ{b}{n, E}
$. Note that the weights $\vectrJ{W}{n, E}\in\mathbb{R}^{D\times M}$ and bias $\vectrJ{b}{n, E}\in\mathbb{R}^{D}$ are per joint $n$.
Following \cite{Vaswani2017Attention}, to inject a notion of ordering we add sinusoidal positional encoding to the joint embeddings, reported to be helpful in extrapolating to longer sequences \cite{Vaswani2017Attention}.
After dropouts \cite{srivastava2014dropout}, the joint embeddings are passed to a stack of $L$ attention blocks where we apply the spatio-temporal attention in parallel to update the embeddings.

\myparagraph{Temporal Attention}
In the temporal attention block, the embedding of every joint is updated by using the past frames of the same joint, i.e., for each joint $n \in \{1 .. N \}$, we calculate a temporal summary $\vectrJ{\Bar{E}}{n} = [\vectrTJ{\Bar{e}}{1}{n}, \dotsc, \vectrTJ{\Bar{e}}{T}{n}]^T \in \mathbb{R}^{T \times D}$.

We use the scaled dot-product attention proposed by \cite{Vaswani2017Attention}, requiring \textit{query} $\vectr{Q}$, \textit{key} $\vectr{K}$, and \textit{value} $\vectr{V}$ representations. Intuitively, the \textit{value} corresponds to the set of past representations that are indexed by the \textit{keys}. For the joint of interest, we compare its \textit{query} representation with all \textit{keys} w.r.t. the dot-product similarity. If the \textit{query} and the \textit{key} are similar (i.e., high attention weight), then the corresponding \textit{value} is assumed relevant. The attention operation yields a weighted sum of \textit{values} $\vectr{V}$:
\begin{align}
\label{eq:attention}
\begin{split}
\text{Attn}(\vectr{Q}, \vectr{K}, \vectr{V}, \vectr{M}) &=  \tau\left(\frac{\vectr{Q} \vectr{K}^T}{\sqrt{D}} + \vectr{M}\right) \vectr{V} 
   = \vectr{A} \vectr{V}
\end{split}
\end{align}
where the mask $\vectr{M}$ prevents information leakage from future steps and $\tau$ is either the softmax function $\sigma$ or a simple normalization by the sum of all attention scores. For the temporal attention mechanism, we refer to matrix $\vectr{A}$ as $\vectr{\Bar{A}} \in \mathbb{R}^{T \times T}$. It contains the temporal attention weights, where each row $i$ in $\vectr{\Bar{A}}$ represents how much attention is given to the previous frames in the sequence. 

The $\vectr{Q}, \vectr{K}$, and  $\vectr{V}$ matrices are projections of the input embeddings $\vectrJ{E}{n} = [\vectrTJ{e}{1}{n}, \dotsc, \vectrTJ{e}{T}{n}]^T \in \mathbb{R}^{T \times D}$. Following \cite{Vaswani2017Attention}, we use a multi-head attention (MHA) mechanism to project the $D-$dimensional representation into subspaces calculated by different attention heads $i \in \{1 .. H\}$:
\begin{align}
\begin{split}
\vectrJ{Q}{n, i} &= \vectrJ{E}{n}\vectrJ{W}{n, Q, i}, \ \ \vectrJ{Q}{n, i} \in \mathbb{R}^{T \times F} \\
\vectrJ{K}{n, i} &= \vectrJ{E}{n}\vectrJ{W}{n, K, i}, \ \ \vectrJ{K}{n, i} \in \mathbb{R}^{T \times F} \\ 
\vectrJ{V}{n, i} &= \vectrJ{E}{n}\vectrJ{W}{n, V, i}, \ \ \vectrJ{V}{n, i} \in \mathbb{R}^{T \times F}
\end{split}
\end{align}
where we set $F = D/H$. Using multiple heads allows the model to gather information from different sets of timesteps into a single embedding. For  example, every head in a MHA with 4 heads, outputs $16$-dimensional chunks of a $64$-dimensional representation. Hence, different attention heads enable accessing different components. The results are then concatenated and projected back into representation space using the weight matrix $\vectrJ{W}{n, O} \in \mathbb{R}^{HF \times D}$ 
\begin{align}
\begin{split}
\text{head}_i &= \text{Attn}\left(\vectrJ{Q}{n, i}, \vectrJ{K}{n, i}, \vectrJ{V}{n, i}, \vectr{M}\right) \\
\vectrJ{\Bar{E}}{n} &= \text{Concat}\left(\text{head}_1, \dotsc, \text{head}_H\right)\vectrJ{W}{n, O} .
\end{split}
\end{align}

\begin{figure*}[t]
    \centering
    \includegraphics[width=\textwidth]{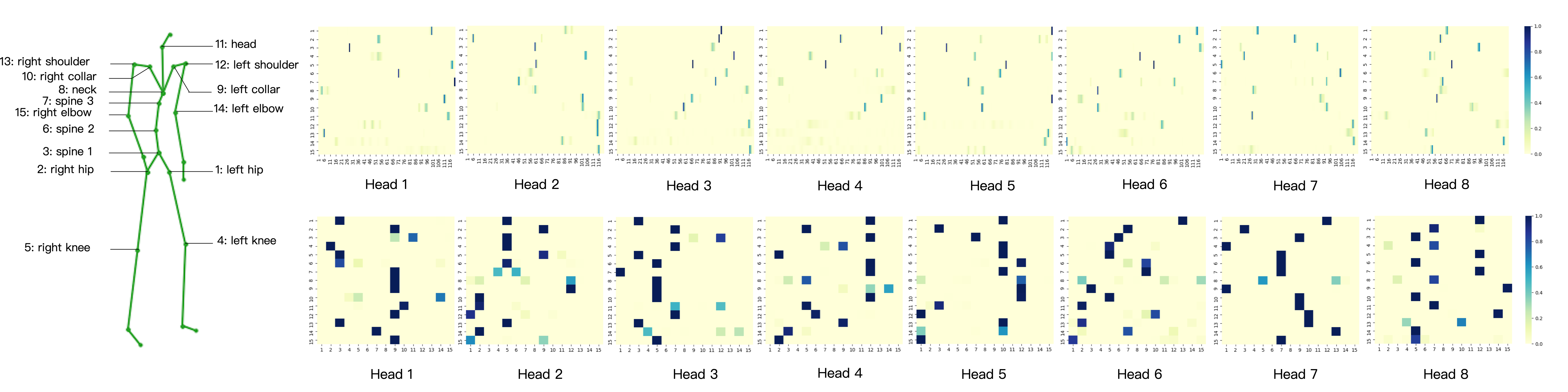}
    \caption{Temporal (\textit{top}) and spatial (\textit{bottom}) attention weights of the first layer given $120$ frames. Each row corresponds to a joint's attention pattern. The columns represent time steps (\textit{top}) or joints (\textit{bottom}). We visualize $8$ attention heads showing that they attend to different joints and time steps. Similarly, different joints exhibit different spatial and temporal attention patterns.}
    \label{fig:firstlayer}
\end{figure*}

\myparagraph{Spatial Attention}
In the vanilla Transformer, the attention block operates on the entire input vector $\vectr{x}_t$ and the relation between the elements are implicitly captured. We introduce an additional spatial attention block to learn dynamics and inter-joint dependencies from the data explicitly. %
The spatial attention mechanism considers all joints of the same timestep. Moreover, the projections we use to calculate the \textit{key} and \textit{value} are shared across joints. Since we aim to identify the most relevant joints, we project them into the same embedding space and compare with the joint of interest. 

For a given pose embedding $\vectrT{E}{t} = [\vectrTJ{e}{t}{1}, \dotsc, \vectrTJ{e}{t}{N}]^T \in \mathbb{R}^{N \times D}$, the spatial summary of joints $\vectrTJ{\tilde{E}}{t}{n}$ is calculated as a function of all other joints by using the multi-head attention:
\begin{align}
\vectrTJ{Q}{t}{i} &= \left[\left(\vectrJ{W}{1, Q, i}\right)^T\vectrTJ{e}{t}{1},  \ldots, \left(\vectrJ{W}{N, Q, i}\right)^T\vectrTJ{e}{t}{N}\right]^T \nonumber\\ 
\vectrTJ{K}{t}{i} &= \vectrT{E}{t}\vectrJ{W}{K, i}, \quad
\vectrTJ{V}{t}{i} = \vectrT{E}{t}\vectrJ{W}{V, i} \\ \nonumber
\text{head}_i &= \text{Attn}(\vectrTJ{Q}{t}{i}, \vectrTJ{K}{t}{i}, \vectrTJ{V}{t}{i}, \vectr{0}) = \vectr{\tilde{A}} \vectrTJ{V}{t}{i}\\ \nonumber
\vectrTJ{\tilde{E}}{t}{n} &= \text{Concat}\left(\text{head}_1, \dotsc,  \text{head}_H\right)\vectrJ{W}{O} \nonumber
\end{align}
where $\vectrTJ{Q}{t}{i} \in \mathbb{R}^{N \times S}$, $\vectrTJ{K}{t}{i} \in \mathbb{R}^{N \times S}$, $\vectrTJ{V}{t}{i} \in \mathbb{R}^{N \times S}$, $S = F = D/H$, $\text{head}_i \in \mathbb{R}^{N \times S}$, and $\vectrJ{W}{O} \in \mathbb{R}^{HS \times D}$. The spatial attention $\vectr{\tilde{A}} \in \mathbb{R}^{N \times N}$ denotes how much attention a joint $i$ pays to the other joints $j$. Since it iterates over the joints at a single time-step, we no longer require a mask.

\myparagraph{Aggregation}
The temporal and spatial attention blocks run in parallel to calculate summaries $\vectr{\Bar{E}}$ and $\vectr{\tilde{E}}$, respectively. They are summed and passed to a $2$-layer pointwise feedforward network \cite{Vaswani2017Attention}, which is followed by a dropout and layer normalization. We stack $L=8$ such attention layers to successively update the joint embeddings and thus to refine the pose predictions. %

\myparagraph{Joint Predictions}
Finally, the joint prediction $\vectrTJ{\hat{j}}{t+1}{n}$ is obtained by projecting the corresponding $D$-dimensional embedding $\vectrTJ{e}{t}{n}$ from the $L$-th attention layer back to the $M$-dimensional joint angle space. Like \cite{Martinez2017Motion}, we apply a residual connection between the previous pose and the prediction.

\subsection{Training and Inference}
We train our model by predicting the next step both for the seed and the target sequences. We use the per-joint $\ell_2$ distance on rotation matrices directly \cite{Aksan_2019_ICCV}:
$$
\mathcal{L}(\vectr{X}, \vectr{\hat{X}}) = \sum\limits_{t=2}^{T+1}\sum\limits_{n=1}^N \norm{\vectrTJ{j}{t}{n} - \vectrTJ{\hat{j}}{t}{n}}_2
$$

At test time, we compute the prediction in an auto-regressive manner. That is, given a pose sequence $\{\vectrT{x}{1}, \dotsc, \vectrT{x}{T}\}$, we get the prediction $\vectrT{\hat{x}}{T+1}$. Due to memory limitations, we apply the temporal attention over a sliding window of poses which we set as the length of the seed sequence. In other words, to produce $\vectrT{\hat{x}}{T+2}$ we condition on the sequence $\{\vectrT{x}{2}, \dotsc, \vectrT{\hat{x}}{T+1}\}$.

%% file: chapters/4_experiments.tex

\begin{table*}
    \centering
    \caption{\textbf{AMASS results}. Lower is better for \textit{Euler}, \textit{Joint Angle} and \textit{Positional} metrics. For the \textit{Area Under the Curve (AUC)}, higher is better. Column-wise best result in \textbf{bold}, second best \underline{underlined}. In contrast to \cite{Aksan_2019_ICCV} we report the mean over the sequence, rather than the sum, and the positional metric was converted from meters to millimeters. ST-Transformer stands for our Spatio-temporal Transformer model. * indicates our own evaluation of the respective model after hyper-parameter tuning.}
    \vspace{-0.5em}
    \resizebox{\linewidth}{!}{%
    \begin{tabular} {l  c c c c | c c c c | c c c c | c c c c }
		& \multicolumn{4}{c}{Euler $\downarrow$} & \multicolumn{4}{c}{Joint Angle $\downarrow$} & \multicolumn{4}{c}{Positional $\downarrow$} & \multicolumn{4}{c}{PCK (AUC) $\uparrow$} \\       
		milliseconds & 100 & 200 & 300 & 400 & 100 & 200 & 300 & 400 & 100 & 200 & 300 & 400 & 100 & 200 & 300 & 400 \\
		\hline
		
		Zero-Velocity \cite{Martinez2017Motion, Aksan_2019_ICCV} & 0.318 & 0.494 & 0.631 & 0.741 & 0.061 & 0.102 & 0.136 & 0.164 & 23.6 & 39.6 & 53.1 & 64.2 & 86.4 & 83.3 & 84.5 & 81.8\\
		
		Seq2seq \cite{Martinez2017Motion,Aksan_2019_ICCV} & 0.336 & 0.499 & 0.623 & 0.722 & 0.062 & 0.097 & 0.126 & 0.150 & 23.6 & 37.4 & 48.6 & 57.9 & 86.5 & 83.8 & 85.4 & 82.9\\
		
		QuaterNet \cite{Pavllo2018,Aksan_2019_ICCV} & 0.248 & 0.391 & 0.509 & 0.606 & 0.044 & 0.074 & 0.102 & 0.125 & 16.7 & 28.7 & 39.6 & 49.0 & 90.4 & 87.4 & 88.0 & 85.4\\
		
		RNN-SPL \cite{Aksan_2019_ICCV} & 0.221 & 0.344 & 0.446 & 0.535 & 0.037 & 0.061 & 0.084 & 0.105 & 13.6 & 23.0 & 31.9 & 40.0 & 92.5 & 89.9 & 90.3 & 87.9\\
		\hline
		
		LTD-10-10* \cite{Mao_2019_ICCV} & 0.205 & 0.333 & 0.447 & 0.544 & 0.039 & 0.065 & 0.089 & 0.111 & 15.6 & 25.7 & 35.3 & 44.1 & 91.5 & 88.8 & 89.3 & 86.9\\
		LTD-Attention* \cite{wei2020his} & 0.207 & 0.321 & 0.418 & \underline{0.499} & 0.035 & 0.060 & 0.083 & \underline{0.102} & 13.4 & 23.2 & 32.1 & \underline{39.7} & 92.5 & 89.8 & 90.2 & \underline{87.9}\\
		
		\hline
		Transformer & 0.216 & 0.332 & 0.433 & 0.523 & 0.036 & 0.060 & 0.083 & 0.104 & 13.3 & 22.8 & 31.8 & 40.0 & 92.5 & 89.9 & 90.2 & 87.8\\
		ST-Transformer & \textbf{0.178} & \textbf{0.291} & \textbf{0.395} & \textbf{0.490} & \underline{0.033} & \underline{0.057} & \underline{0.081} & 0.103 & \underline{13.2} & \underline{22.6} & \underline{32.0} & 40.9 & \underline{93.1} & \underline{90.2} & \underline{90.3} & 87.7\\
		ST-Transformer w/o $\sigma$ & \underline{0.187} & \underline{0.302} & \underline{0.409} & 0.504 & \textbf{0.032} & \textbf{0.056} & \textbf{0.079} & \textbf{0.101} & \textbf{12.8} & \textbf{21.8} & \textbf{30.9} & \textbf{39.5} & \textbf{93.5} & \textbf{90.6} & \textbf{90.7} & \textbf{88.2}\\
		\hline
	\end{tabular}
    }
    \label{tab:amass_new}
\end{table*}

\section{Experiments}
\label{sec:experiments}
We evaluate the ST-Transformer on AMASS \cite{AMASS2019} and H3.6M \cite{h36m} in \secref{sec:eval_quant} following the standard protocols for short-term predictions and adopting distribution-based metrics for long-term predictions. We note that both benchmarks focus on modeling 3D joint angles unlike the 3D position-based benchmarks presented in the previous work \cite{Wang_2019_ICCV, wei2020his}. We argue that modeling the 3D position representation of human pose is prone to errors as the models are free to violate the skeletal configuration. In other words, the outputs may contain artifacts such as inconsistent bone lengths across the frames \cite{Holden2016, kaufmann2020convolutional}. In contrast, the joint angle representation implicitly preserves the skeletal structure which is important in many downstream tasks.

\secref{sec:eval_qual} and \secref{sec:eval_dissection} show qualitative results and attention weights, thus providing insights into how the model forms predictions.
We validate design choices through ablation studies in \secref{sec:eval_ablation}.
We run our models and the baselines on various joint angle representations including rotation matrix, quaternion and angle-axis, and report the best performance. Implementation details for our model and the baselines are provided in the supplementary material. %

\subsection{Quantitative Evaluation}
\label{sec:eval_quant}

\begin{figure*}
    \centering
        \includegraphics[width=\textwidth]{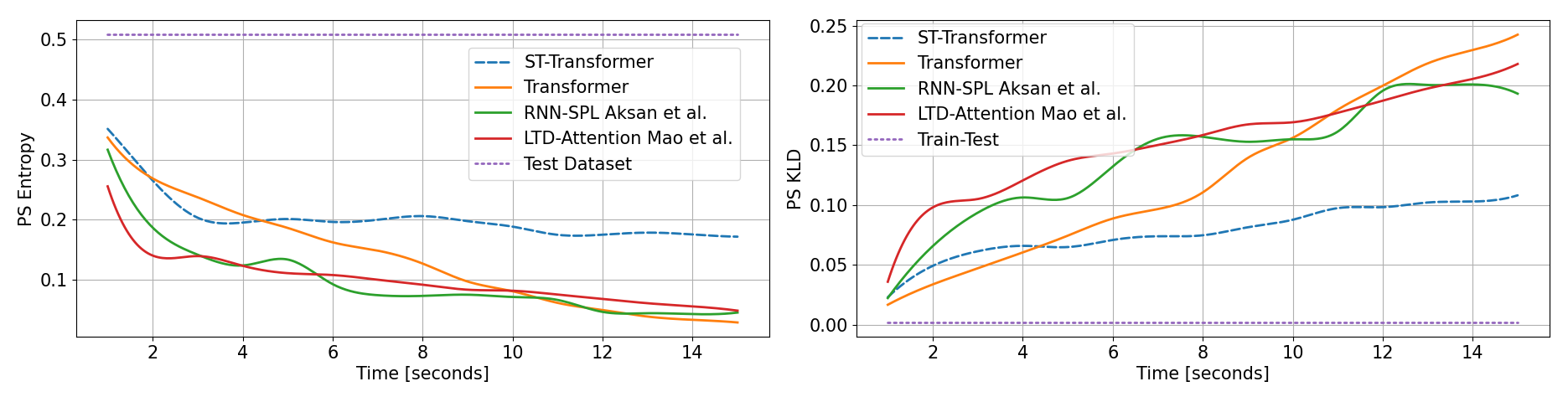}
        \vspace{-1em}
        \caption{\textbf{Power Spectrum (PS) metrics.}
        \textit{(left)} PS Entropy, higher is better. Our model (blue, dashed) exhibits higher entropy suffering less from the static pose problem in longer predictions. \textit{(right)} PS KLD, lower is better. Our model's prediction distribution stays closer to the data distribution as indicated by the symmetric KLD.}
    \label{fig:ps_all}
\end{figure*}

\myparagraph{AMASS}
We follow Aksan \etal \cite{Aksan_2019_ICCV} and evaluate our model on the large-scale motion dataset AMASS \cite{AMASS2019}.
Table \ref{tab:amass_new} summarizes the results with pairwise angle- and position metrics up to $400$ ms. For longer time horizons, direct comparison to the ground-truth via MSE becomes increasingly problematic particularly for auto-regressive models \cite{Fragkiadaki2015ERD}.
Hence, in addition to the standard metrics in \cite{Aksan_2019_ICCV, Martinez2017Motion}, we conduct further analysis by using complementary metrics in the frequency domain allowing benchmarking up to $15$ seconds (\figref{fig:ps_all}).

In the short-term evaluations, we compare our ST-Transformer with the vanilla Transformer, previously reported RNN-based architectures and two DCT-based architectures \cite{Mao_2019_ICCV, wei2020his}. We could not compare to \cite{cai2020learning} as no implementation is publicly available.
The vanilla Transformer follows an auto-regressive approach similar to our ST-Transformer but applies 1D attention on the pose vectors.
Furthermore, we include results of a variant of our model that does not use softmax $\sigma$ in the attention (cf. \eqnref{eq:attention}) as the softmax may lead to gradient instabilities. Instead, we normalize by the sum of all attention scores similar to \cite{wei2020his}. Our ST-Transformers achieve state-of-the-art in all metrics while LTD-Attention \cite{wei2020his} remains competitive at 400 ms.

\myparagraph{Long-Term}
Our approach is fully generative, so while it maintains local consistency, the global positions may deviate from the ground-truth under natural variation. Hence, we propose to use distribution-based metrics in the power spectrum (PS) space proposed by Hernandez \etal \cite{Hernandez_2019_ICCV} instead of a direct comparison with the ground-truth frames.
We report two metrics, (i) \textit{PS KLD} which measures the discrepancy between the prediction and the test distributions via the KL divergence, and (ii) \textit{PS Entropy} capturing the entropy of the prediction distribution in the power spectrum. The latter measures how diverse the predictions are. Models collapsing to static pose predictions end up with lower entropy values. However, \textit{PS Entropy} can be deceived by random predictions and it should thus be interpreted in conjunction with the \textit{PS KLD} where only a set of predictions that are similar to the real data samples can achieve a lower score. 

%

\begin{table*}[h]
    \center
	\caption{\textbf{H3.6M results} for \textit{walking}, \textit{eating}, \textit{smoking} and \textit{discussion} activities. Values are the Euler angle metric measured at the given time step (lower is better). ST-Transformer stands for our Spatio-temporal Transformer model, consistently outperforming the vanilla Transformer and surpassing or performing on par with state-of-the-art.}
	\vspace{-1em}
	\resizebox{\linewidth}{!}{%
	\begin{tabular} {l  c c c c | c c c c | c c c c | c c c c }
		& \multicolumn{4}{c}{Walking} & \multicolumn{4}{c}{Eating} & \multicolumn{4}{c}{Smoking} & \multicolumn{4}{c}{Discussion} \\       
		milliseconds & 80 & 160 & 320 & 400 & 80 & 160 & 320 & 400 & 80 & 160 & 320 & 400 & 80 & 160 & 320 & 400 \\
		\hline
		
		RNN-SPL \cite{Aksan_2019_ICCV} & 0.26 & 0.40 & 0.67 & 0.78 & 0.21 & 0.34 & 0.55 & 0.69 & 0.26 & 0.48 & 0.96 & 0.94 & 0.30 & 0.66 & 0.95 & 1.05\\
		
		AGED \cite{Wang2018Adversarial} & 0.22 & 0.36 & 0.55 & 0.67 & 0.17 & 0.28 & 0.51 & 0.64 & 0.27 & 0.43 & \textbf{0.82} & 0.84 & 0.27 & 0.56 & 0.76 & 0.83  \\
	    
		\hline
		LTD-10-10 \cite{Mao_2019_ICCV} & 0.18 & 0.31 & 0.49 & \textbf{0.56} & 0.16 & 0.29 & 0.50 & 0.62 & 0.22 & 0.41 & 0.86 & 0.80 & 0.20 & 0.51 & 0.77 & 0.85\\
		
		LTD-Attention \cite{wei2020his} & 0.18 & \textbf{0.30} & \textbf{0.46} & 0.51 & 0.16 & 0.29 & 0.49 & 0.60 & 0.22 & 0.42 & 0.86 & 0.80 & 0.20 & 0.52 & 0.78 & 0.87\\
		
		Propagation \cite{cai2020learning} & \textbf{0.17} & \textbf{0.30} & 0.51 & 0.55 & 0.16 & 0.29 & 0.50 & 0.61 & \textbf{0.21} & \textbf{0.40} & 0.85 & \textbf{0.78} & 0.22 & \textbf{0.39} & \textbf{0.62} & \textbf{0.69}\\
		
		\hline
		
		Transformer & 0.25 &	0.42&	0.67&	0.79&	0.21&	0.32&	0.54&	0.68&	0.26&	0.49&	0.94&	0.90&	0.31&	0.67&	0.95&	1.04\\
		
		ST-Transformer & 0.19 & 0.33 & 0.56 & 0.64 & \textbf{0.15} & \textbf{0.27} & \textbf{0.45} & \textbf{0.57} & 0.22 & 0.41 & 0.87 & 0.82 & \textbf{0.19} & 0.53 & 0.81 & 0.93\\
		
		\hline
	\end{tabular}
	}
	\label{tab:h36m_new}
\end{table*}

In Figure \ref{fig:ps_all}, we compare our ST-Transformer, the vanilla Transformer, RNN-SPL \cite{Aksan_2019_ICCV} and LTD-Attention with DCT representations \cite{wei2020his} on these PS metrics for predictions up to $15$ seconds. To produce long sequences with LTD-Attention, we run it autoregressively on its own predictions and in a sliding window fashion.
The reference values (dotted, purple) from the training and test samples are calculated on randomly extracted 1-second windows (60 frames).
Similarly, we compute statistics over the predictions of the respective model by shifting a 1-second window. Thus, we compare every second of the prediction with real 1-second clips.
As is expected, the prediction statistics do deviate from the ground-truth statistics with increasing prediction horizon.
However, our model remains much closer to the real data statistics than any of the baselines. This indicates that our model does generate plausible poses that are similar to the data distribution without memorizing the exact sequences. 

The PS Entropy plot in \figref{fig:ps_all} shows that the ST-Transformer has a higher entropy than the baselines, thus indicating its power to mitigate the collapse to a static pose.
It furthermore indicates that none of the baselines can alleviate this problem as much as the ST-Transformer. The difference is more pronounced with horizon length.
These observations are additionally corroborated by our visualizations in the supplementary video and \figref{fig:qualitative} where we clearly see that the walking motion produced by the baselines phase out earlier compared to our ST-Transformer's output.

\myparagraph{H3.6M}
Traditionally, motion prediction has been benchmarked on H3.6M \cite{h36m}.
\tabref{tab:h36m_new} compares our model in this setting, where we are competitive and often achieve state-of-the-art.
H3.6M is roughly 14 times smaller than AMASS and its test split consists only of a few sequences, which has been reported to cause high variance \cite{Aksan_2019_ICCV, Pavllo2019Arxiv}.
Furthermore, as is evident from \tabref{tab:h36m_new}, improvements are often marginal and recent works seem to converge to the same error for most actions. 
For these reasons we argue that the AMASS benchmark introduced by \cite{Aksan_2019_ICCV} carries more weight. %

\myparagraph{Discussion}
It is evident that the spatio-temporal decoupling of attention is indeed beneficial when compared to the vanilla Transformer. In all settings our ST-Transformer significantly outperforms the vanilla counterpart. Compared to the RNN-based autoregressive baselines such as RNN-SPL, Seq2seq or AGED, our model makes more accurate predictions in short-term horizon as well as plausible longer-term generations by mitigating the error accumulation problem. 

The DCT-based baselines are the most competitive and also conceptually more relevant to our work. We argue that the task favors the DCT-based representations as a temporal window is encoded and decoded in one go. In other words, the entire prediction horizon is predicted at once in contrast to our frame-by-frame predictions. Hence, we observe that our model shows strong performance in the shortest prediction horizon with respect to the \emph{pairwise comparisons with the ground-truth} on both AMASS and H3.6M (cf. \tabref{tab:amass_new}, \ref{tab:h36m_new}). Our model's error with respect to the ground-truths seems to increase with longer prediction horizons, which is expected for an auto-regressive model due to error accumulation. Yet, it remains very competitive and the distribution-based metrics in Figure \ref{fig:ps_all} also highlight that our model is statistically closer to the real data in very long-term predictions.

\subsection{Qualitative Evaluation}
\label{sec:eval_qual}
Here, we evaluate the generative capabilities up to $20$ seconds. We feed the model with a particular motion sequence of $2$ seconds and auto-regressively predict beyond its training horizon (i.e., $400$ ms or $1$ sec). 

We qualitatively compare our model with the vanilla Transformer, RNN-SPL \cite{Aksan_2019_ICCV} and LTD-Attention \cite{wei2020his} on a walking sample from AMASS in \figref{fig:qualitative}.
With RNN-SPL any variation quickly disappears within 5 seconds.
The vanilla Transformer does not reach a static pose for longer, which shows the benefits of the attention mechanism over recurrent networks.
However, it still collapses around second $15$, whereas our ST-Transformer maintains the walking motion over the entire duration of $20$ seconds.
Also the LTD-Attention model produces walking motion longer than RNN-SPL, but still converges before 10 seconds.
More samples can be found in the appendix and video.

While our model performs well on periodic motions for long horizons, the prediction horizon is limited to a few seconds for aperiodic motion types as the motion cycle is completed.
This still exceeds previously reported horizons significantly.
Also, it is not unexpected since the model is unlikely to be exposed to transition patterns as it is trained on rather short $2$-second windows. %

\begin{table}[b]
    \centering
    \caption{\textbf{2D Attention Ablation} on AMASS.}
    \vspace{-0.5em}
    \resizebox{\linewidth}{!}{%
    \begin{tabular} {l  c c | c c | c c }
		& \multicolumn{2}{c}{Euler $\downarrow$} & \multicolumn{2}{c}{Joint Angle $\downarrow$} & \multicolumn{2}{c}{Positional $\downarrow$} \\       
		milliseconds & 100 & 400 & 100 & 400 & 100 & 400 \\
		\hline
		2D-Transformer & 0.213 & 0.555 & 0.037 & 0.115 & 14.5 & 45.2\\
		ST-Transformer & \textbf{0.178} & \textbf{0.490} & \textbf{0.033} & \textbf{0.103} & \textbf{13.2} & \textbf{39.5}\\
		\hline
	\end{tabular}
    }
    \label{tab:2d_ablation}
\end{table}
\begin{figure}[b]
    \centering
    \includegraphics[width=1.0\columnwidth]{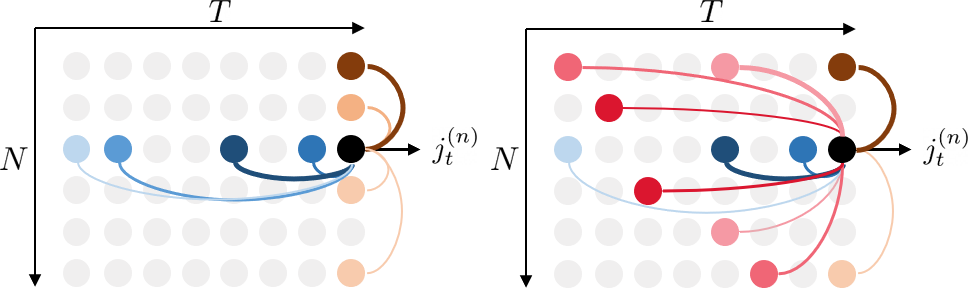}
    \caption{\textbf{Spatio-temporal attention vs. 2D attention.} All the joints except the future ones are available in the naive 2D attention case as highlighted in red-ish colors.}
    \label{fig:2d_attention}
\end{figure}

\begin{figure*}[t]
    \centering
    \includegraphics[width=\textwidth]{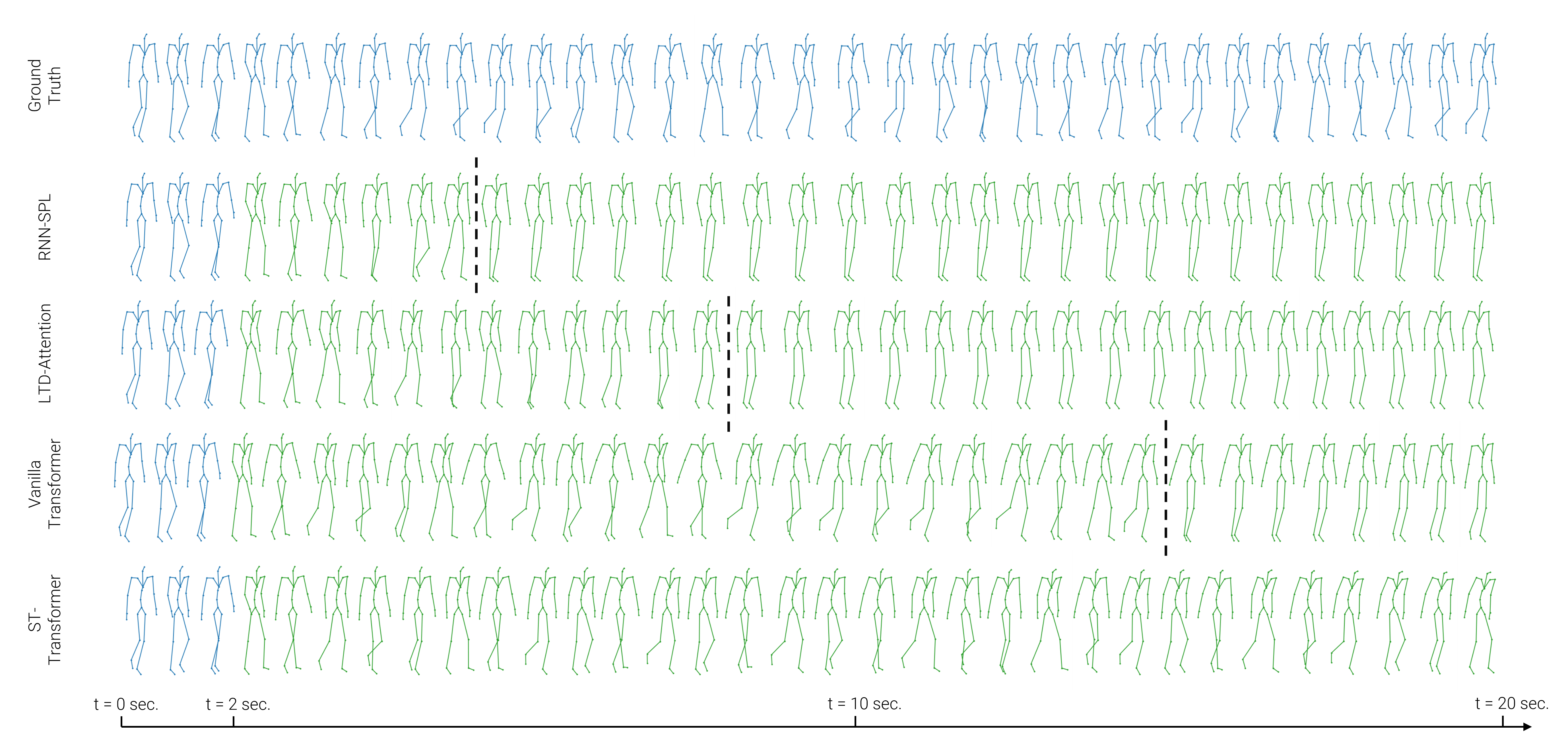}
    \caption{\textbf{Qualitative Results.} A walking motion predicted up to 20 seconds by RNN-SPL \cite{Aksan_2019_ICCV}, LTD-Attention \cite{wei2020his}, vanilla Transformer, and our ST-Transformer. Seed length is 2 sec. The dashed lines indicate when the motion comes to a halt.}
    \label{fig:qualitative}
\end{figure*}
\subsection{Ablations}
\label{sec:eval_ablation}
\myparagraph{2D Attention}
To unpack our contribution more clearly we implement and compare to a 2D transformer architecture. Here, every joint attends to all other joints across all frames (\figref{fig:2d_attention}). The complexity of the attention thus becomes $O(N \times T)$. With this design, memory requirements increase drastically. Hence, for training we either reduce the model complexity or decrease $T$ or the batch size. The performance of the best configuration we found is summarized in \tabref{tab:2d_ablation}. It clearly falls behind the ST-Transformer. Our approach reduces the complexity from $O(N \times T)$ to $O(N+T)$, allowing to attend to a longer history and to use a larger model, highlighting our contribution on an architectural level.

\begin{figure}[b]
    \centering
    \includegraphics[width=0.75\columnwidth]{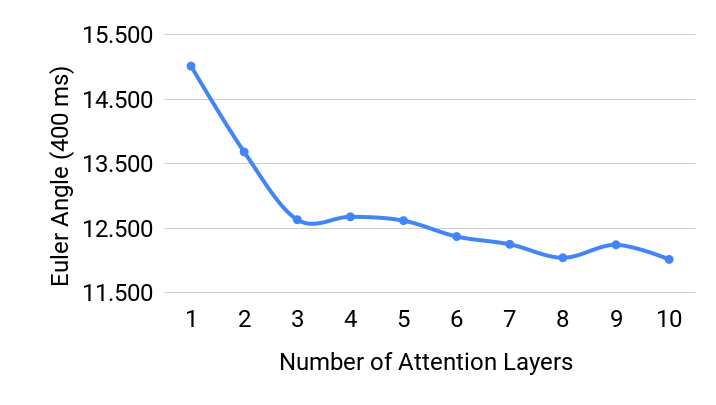}
    \vspace{-1em}
    \caption{\textbf{Ablation} on value of $L$ on AMASS.}
    \label{fig:ablation_layers}
\end{figure}

\myparagraph{Number of Layers}
We train our model with varying number of attention layers $L$. \figref{fig:ablation_layers} shows that competitive performance on AMASS is reached with only $3$ layers. As the number of layer increases, the representations learned by our model are tuned better. It can be considered as the number of message passing steps to update the available representation.

\subsection{What does Attention Look Like?}
\label{sec:eval_dissection}
The underlying attention mechanism of our model provides insights into the model's predictions. \figref{fig:firstlayer} visualizes the temporal and spatial attention weights $\vectr{\Bar{A}}$, $\vectr{\tilde{A}}$ taken from an AMASS sequence. Those weights are used to predict the first frame given the seed sequence of $120$ frames. In our visualizations, we used the first layer as it uses the initial joint embeddings directly. In the upper layers, the information is already gathered and hence the attention is dispersed.

First, we observe that there is a diverse set of attention patterns across heads, enriching the representation through gathering information from multiple sources. Second, the temporal attention weights reveal that the model is able to attend over a long horizon. While some heads focus on near frames, some look to the very beginning which would be difficult for an RNN. Finally, in the spatial attention, we observe joint-dependencies not only on the kinematic chain but also across the left and right parts of the skeleton. For example, while predicting the left knee joint, the model attends to the spine, left collar and right hip, knee and collar joints. Similarly, for the right elbow, the most informative joints are spine, the left collar, the hips and knee joints. To see how attention weights change over time, 
please refer to 
Fig. 7 in the appendix and the video.

%% file: chapters/8_conclusion.tex

\section{Conclusion}
We introduce a novel spatio-temporal transformer (ST-Transformer) network for generative modeling of 3D human motion.
Our proposed architecture learns intra- and inter-joint dependencies explicitly via its decoupled temporal and spatial attention blocks.
We show that the self-attention concept can be very effective in learning representations for both short- and long-term motion predictions compared to the DCT-based motion representations. Furthermore, it mitigates the long-term dependency issue observed in auto-regressive architectures and is able to synthesize motion sequences up to $20$ seconds conditioned on periodic motion types such as locomotion.